\relax
\documentclass[letterpaper]{article}
\usepackage{acl2012}
\usepackage{times}
\usepackage{helvet}
\usepackage{courier}
\usepackage{amsmath}
\usepackage{amsfonts}
\usepackage{graphicx}
\usepackage{subcaption}

\DeclareMathOperator{\sigmoid}{sigmoid}

\makeatletter
\newcommand{\@BIBLABEL}{\@emptybiblabel}
\newcommand{\@emptybiblabel}[1]{}
\makeatother

\begin{document}
%
\title{A Dynamic Window Neural Network for CCG Supertagging}
\newcommand*{\affaddr}[1]{#1} 
\newcommand*{\affmark}[1][*]{\textsuperscript{#1}}
\newcommand*{\email}[1]{\texttt{#1}}

\author{%
Huijia Wu\affmark[1,3], Jiajun Zhang\affmark[1,3], and Chengqing Zong\affmark[1,2,3]\\
\affaddr{\affmark[1]National Laboratory of Pattern Recognition, Institute of Automation, CAS}\\
\affaddr{\affmark[2]CAS Center for Excellence in Brain Science and Intelligence Technology }\\
\affaddr{\affmark[3]University of Chinese Academy of Sciences} \\
\email{\{huijia.wu,jjzhang,cqzong\}@nlpr.ia.ac.cn}
}
\maketitle

\begin{abstract}
Combinatory Category Grammar (CCG) supertagging is a task to assign lexical categories to each word in a sentence. Almost all previous methods use fixed context window sizes as input features. However, it is obvious that different tags usually rely on different context window sizes. These motivate us to build a supertagger with a dynamic window approach, which can be treated as an attention mechanism on the local contexts. Applying dropout on the dynamic filters can be seen as drop on words directly, which is superior to the regular dropout on word embeddings. We use this approach to demonstrate the state-of-the-art CCG supertagging performance on the standard test set. 
\end{abstract}

\section{Introduction}
Combinatory Category Grammar (CCG) provides a connection between syntax and semantics of natural language. The syntax can be specified by derivations of the lexicon based on the combinatory rules, and the semantics can be recovered from a set of predicate-argument relations. CCG provides an elegant solution for a wide range of semantic analysis, such as semantic parsing \cite{zettlemoyer2007online,kwiatkowski2010inducing,kwiatkowski2011lexical,artzi-lee-zettlemoyer:2015:EMNLP}, semantic representations \cite{bos2004wide,bos2005towards,bos2008wide,lewis2013combining}, and semantic compositions, all of which heavily depend on the supertagging and parsing performance. All these motivate us to build a more accurate CCG supertagger.

CCG supertagging is the task to predict the lexical categories for each word in a sentence. Existing algorithms on CCG supertagging range from point estimation \cite{clark2007wide,lewis2014improved} to sequential estimation \cite{xu2015ccg,lewis2016lstm,vaswani2016supertagging}, which predict the most probable supertag of the current word according to the context in a fixed size window. This fixed size window assumption is too strong to generalize. We argue this from two perspectives.

One perspective comes from the inputs. For a particular word, the number of its categories may vary from 1 to 130 in CCGBank 02-21 \cite{hockenmaier2007ccgbank}. We need to choose different context window sizes to meet different ambiguity levels. The other perspective is for the targets. There are about 21000 different words together with 31 different Part-Of-Speech(POS) tags which have the same category $N/N$. Using the same context window size for each word is obviously inappropriate.

\begin{figure}
\centering
\includegraphics[width=1\linewidth, scale=1]{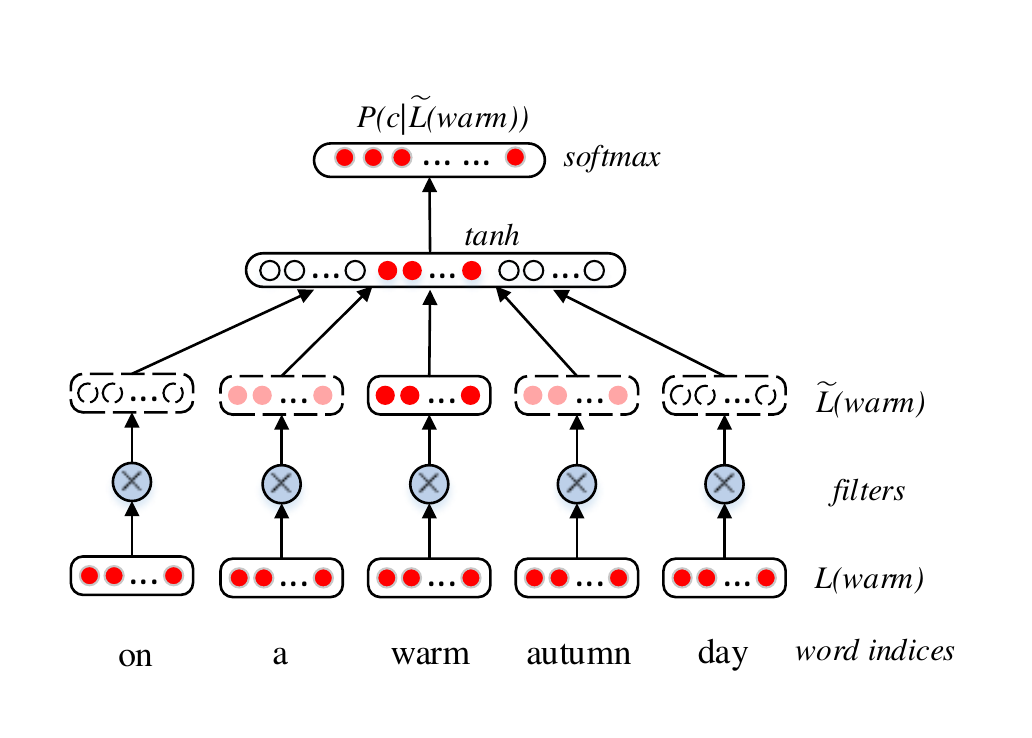}
\caption{A dynamic window approach for supertagging using a multilayer perceptron.}
\label{fig:mlp}
\end{figure}

To overcome these problems, we notice that Xu et al. \shortcite{xu2015ccg} use dropout in the embedding layer to make the input contexts sparse. This method motivates us to get rid of the unnecessary information in the contexts automatically rather than use a pre-specified prior. Then we observe that the gating mechanism of long short-term memory (LSTM) blocks, especially the input gate, can determine when to enter into the block. All these inspire us to add a gate to each item in the context windows to make them sparse but informative.

This method is naturally an extension to the encoder-decoder with the attention mechanism \cite{bahdanau2014neural}, which can be interpreted as focusing on parts of the memories when making decisions. From this perspective, the contexts of the current word are the memories, and the dynamic window is the attention. We focus on the contexts extracted from the attended windows to predict the corresponding lexical categories.

Figure \ref{fig:mlp} visualize this method. We add one logistic gate to each item in the input contexts. If the gate is close to zero, the corresponding features in the window will be ignored during training. Moreover, we add a dropout mask on the gates to further improve its sparsity, which can be treated as a word-level dropout. Combining attention and dropout lead a significant performance improvement. 

We evaluated our approach on multilayer perceptrons (MLPs) and and recurrent neural networks (RNNs), including vanilla forms (standard RNNs) and gated RNNs. The experiments show that the performance of these networks can obtain improvements using our method, and can outperform all the reported results to date.

\section{Background}

\subsection{Category Notation}

CCG uses a set of lexical categories to represent constituents \cite{steedman2000syntactic}. In particular, a fixed finite set is used as a basis for constructing other categories, 
which is described in Table \ref{basic_cate}.

\begin{table}[h]
\begin{center}
\begin{tabular}{c|c}
\hline Category & Description \\ \hline \hline
N & noun \\ \hline
NP & noun phrase \\ \hline
PP & prepositional phrase \\ \hline
S & sentence \\
\hline
\end{tabular}
\end{center}
\caption{\label{basic_cate} The description of basic categories used in CCG}
\end{table}

The basic categories could be used to generate an infinite set $\mathbb{C}$ of functional categories by applying the following recursive definition:

\begin{itemize}
\item $N, NP, PP, S \in \mathbb{C}$
\item $X/Y, X\backslash Y \in \mathbb{C}$ if $X, Y \in \mathbb{C}$
\end{itemize}

Each functional category specifies some arguments. Combining the arguments can form a new category according to the orders \cite{steedman2011combinatory}. The argument could be either basic or functional, and the orders are determined by the forward slash $/$ and the backward slash $\backslash$. A category $X/Y$ is a forward functor which could accept an argument $Y$ to the right and get $X$, while the backward functor $X\backslash Y$ should accept its argument $Y$ to the left.

\subsection{Neuron Based Supertaggers}
CCG supertagging is an approach for assigning lexical categories for each word in a sentence. The problem can be formulated by $P(\mathbf{c} | \mathbf{w}; \pmb{\theta})$, where $\mathbf{w} = [w_1, \ldots, w_T]$ indicates the $T$ words in a sentence, and $\mathbf{c} = [c_1, \ldots, c_T]$ indicates the corresponding lexical categories. Notice that the length of the words and categories are the same. We denote vectors with bolded font and matrices with capital letters. Bias terms in neural networks are omitted for readability.

\subsubsection{Network Inputs.}
Network inputs are the representation of each token in a sequence. Our inputs include the concatenation of word representations, character representations, and capitalization representations. To reduce sparsity, all words are lower-cased, together with capitalization and character representations as inputs.

Formally, we can represent the distributed word feature $\mathbf{f}_{w_t}$ using a concatenation of these embeddings:
\begin{align}
    \mathbf{f}_{w_t} = [\mathbf{L}_w(w_t); \mathbf{L}_a(a_t); \mathbf{L}_c(\mathbf{c}_w)]
\end{align}
where $w_t$, $a_t$ represent the current word and its capitalization. $\mathbf{c}_w := [c_1, c_2, \ldots, c_{T_w}]$, where $T_w$ is the length of the word and $c_i, i \in \{1, \ldots, T_w\}$ is the $i$-th character for the particular word.  $\mathbf{L}_w(\cdot) \in \mathbb{R}^{|V_w|\times n}$, $\mathbf{L}_a(\cdot) \in \mathbb{R}^{|V_a|\times m}$ and $\mathbf{L}_c(\cdot) \in \mathbb{R}^{|V_c| \times r}$ are the look-up tables for the words, capitalization and characters, respectively. $\mathbf{f}_{w_t} \in \mathbb{R}^{n+m+r}$ represents the distributed feature of $w_t$. A context window of size $d$ surrounding the current word is used as an input:
\begin{align} \label{eq:input}
    \mathbf{x}_t = [\mathbf{f}_{w_{t-\lfloor d/2 \rfloor}}; \ldots; \mathbf{f}_{w_{t+\lfloor d/2 \rfloor}}]
\end{align}
where $\mathbf{x}_t \in \mathbb{R}^{(n+m+r)\times d}$ is the concatenation of the context features. We use it as the input of the network.

\subsection{Network Outputs.}
Since our goal is to assign CCG categories to each word, we use a \emph{softmax} activation function $g(\cdot)$ in the output layer:
\begin{align}
\mathbf{y}_t &= g(\mathbf{W^{hy}} \mathbf{h}_t)
\end{align}
where $\mathbf{y}_t \in \mathbb{R}^{K}$ is a probability distribution over all possible categories. $y_k(t) = \frac{\exp(h_k)}{\sum_{k'} \exp(h_{k'})}$ is the $k$-th dimension of $\mathbf{y}_t$, which corresponds to the $k$-th lexical category in the lexicon. $\mathbf{h}_t \in \mathbb{R}^H$ is the output of the hidden layer at time $t$. $\mathbf{W^{hy}} \in \mathbb{R}^{K \times H}$ is the hidden-to-output weight.

\subsubsection{Inputs to Outputs Mappings.}
Neuron based supertaggers model the inputs to outputs mappings using neural networks. Since CCG supertagging can either be treated as a point or a sequential estimation problem, which correspond to two kinds of neural networks: MLPs and RNNs, respectively. For simplicity, we will talk about gated RNNs only, which are special kinds of RNNs with logistic gates in the hidden units to control information flow. There are many kinds of gated RNNs, such as long short-term memory, \cite{hochreiter1997lstm} and gated recurrent unit \cite{cho2014learning}. We focus on LSTMs only in this work.

LSTMs replace the hidden units in vanilla RNNs with complicated blocks, which are designed as:
\begin{align}
\mathbf{\tilde{c}}_t &= \sigma(\mathbf{W^{xc}}\mathbf{x}_t + \mathbf{W^{hc}}\mathbf{h}_{t-1}) \\
\mathbf{c}_t &= \mathbf{f}_t \odot \mathbf{c}_{t-1} + \mathbf{i}_t \odot \mathbf{\tilde{c}}_t \\
\mathbf{h}_t &= \mathbf{o}_t \odot f(\mathbf{c}_t)
\end{align}
where $\mathbf{x}_t$ and $\mathbf{h}_t$ are the input and the output of the block. $\mathbf{i}_t \in \mathbb{R}^{H}$, $\mathbf{f}_t \in \mathbb{R}^{H}$ and $\mathbf{o}_t \in \mathbb{R}^{H}$ denote the input gate, forget gate and output gate, respectively, which are logistic units to filter the information. $\mathbf{W^{xc}} \in \mathbb{R}^{H \times I}$ is the weight storing the input. $\mathbf{W^{hc}} \in \mathbb{R}^{H \times H}$ is the recurrent weight connecting the previous block outputs to the cells. $\mathbf{c}_t \in \mathbb{R}^{H}$ is the short-term memory state, which is used to store the history information. Based on the three gates, the information flow in $\mathbf{c}_t$ can be kept for a long time. $f(\cdot)$ is the non-linear mapping, here we use the hyperbolic tangent function $f(z) = \frac{e^{z} - e^{-z}}{e^z + e^{-z}}$.

\section{A Dynamic Window Approach}
In this section we will introduce a dynamic window approach for supertagging. Specifically, we add logistic gates to each token in the context window to filter the unnecessary information. We can modify the Eq. \eqref{eq:input} to:

\begin{align} \label{eq:xt}
&\mathbf{\tilde{x}}_t = \mathbf{r}_t \otimes \mathbf{x}_t \\ 
&:= [r_{{t-\lfloor d/2 \rfloor}} \mathbf{f}_{w_{t-\lfloor d/2 \rfloor}}; \ldots  ; r_{{t+\lfloor d/2 \rfloor}}\mathbf{f}_{w_{t+\lfloor d/2 \rfloor}}] \label{eq:tilde-L}
\end{align}

Here $\otimes$ denote an element-wise scalar-vector product. $\mathbf{r}_t := [r_{{t-\lfloor d/2 \rfloor}}, \ldots, r_{{t+\lfloor d/2 \rfloor}}] \in \mathbb{R}^{d}$ is a logistic gate to filter the unnecessary contexts. $r_i \text{ and } \mathbf{f}_i, i \in \{t-\lfloor d/2 \rfloor, \ldots, t+\lfloor d/2 \rfloor\}$ is a scalar and a vector, respectively. Their product is defined as:
\begin{align*}
	r\mathbf{f} := [r f_1, \ldots, r f_n]
\end{align*}

One perspective for the filter gate is an attention model focusing on the necessary contexts. This effect can be visualized in Figure \ref{fig:mlp}: If one component of $\mathbf{r}$, say $r_i$ is 0, the corresponding word feature $\mathbf{f}_{w_i}$ will be removed or deactivated from the input.

\subsection{Design of the Gates}
We use a feed-forward neural network to learn $\mathbf{r}_t$:
\begin{align}
	\mathbf{r}_t &= \sigma(\mathbf{W^{xr}} \mathbf{x}_t) \label{eq:filter-lt}
\end{align}
where $\sigma(\cdot)$ is the $\sigmoid$ function defined as $\sigma(z) = \frac{1}{1 + e^{-z}}$. This function is to make sure the values of $\mathbf{r}_t$ are between 0 and 1. $\mathbf{x}_t$ is network input, as defined in Eq. \eqref{eq:input}. $\mathbf{r}_t \in \mathbb{R}^d$ is the output of the dynamic window model, where $d$ is the window size. $\mathbf{W^{xr}} \in \mathbb{R}^{d \times I}$ is the weight to be learned.

One disadvantage of the $\sigmoid$ function is when a neuron is nearly saturated, its derivative becomes small, which makes the connecting weights change very slowly. If some neurons in $\mathbf{r}_t$ are saturated, which states will be stable and may not generalize well. To further improve the sparsity in the context window, we add a dropout mask \cite{srivastava2014dropout} on $\mathbf{r}_t$:
\begin{align}
    l_i(t) &\sim \text{Bernouli}(p) \\
    \mathbf{\tilde{r}}_t &= \mathbf{l}_t \odot \mathbf{r}_t
\end{align}
where $\mathbf{l}_t$ is a vector of independent Bernouli random variables $l_i(t)$, which has probability $p$ of being 1. Since $\mathbf{\tilde{r}}_t$ acts on each word feature in the context window, this dropout can be viewed as drop on words directly \cite{dai2015semi}. One minor difference is they use word dropout at the sentence level, and we use it at the dynamic window level.

To further explain it, let's consider a trivial case when all items in $\mathbf{r}_t$ are set to 1:
\begin{align} \label{eq:one}
    \mathbf{r}_t = [1, \ldots, 1]
\end{align}

Adding a dropout mask on such a $\mathbf{r}_t$ is equivalent to drop the words in the contexts randomly, which can be seen as a window approach with a random size.

\subsection{Embedded into MLPs}
For MLPs with this approach, we can use the filtered context as an input:
\begin{align}
\mathbf{h}_t &= f(\mathbf{W^{xh}} \mathbf{\tilde{x}}_t) \\
\mathbf{y}_t &= g(\mathbf{W^{hy}} \mathbf{h}_t)
\end{align}
where $\mathbf{\tilde{x}}_t$ is defined in Eq. \eqref{eq:xt}. We use the filtered context as an input to the hidden layer. $\mathbf{W^{xh}} \in \mathbb{R}^{H \times I}$ and $\mathbf{W^{hy}} \in \mathbb{R}^{K \times H}$ is the weight parameters of MLP.

\subsection{Embedded into Vanilla RNNs}
The similar approach can be applied to RNNs with slight modifications. For each hidden state we have two types of inputs: one is from the input layer, the other is from the hidden \cite{elman1990finding} or the output layer \cite{jordan1986attractor}. The recurrent weight may vanish or explode if its eigenvalues are deviated from 1. To avoid these problems, we add one gate to the recurrent input to reset it:
\begin{align}
&\mathbf{\tilde{r}}_t = \sigma(\mathbf{W^{xr}} \mathbf{x}_t) \\
&s_t = \sigma(\mathbf{W^{xs}} \mathbf{x}_t)
\end{align}
where $s_t \in \mathbb{R} $ is a scalar between 0 and 1, which is used to reset the recurrent input. $\mathbf{W^{xs}} \in \mathbb{R}^{1 \times I}$ is the corresponding hidden-to-output weight. Intuitively, if $s_t$ is close to zero, the recurrent input $\mathbf{W^{yc}}\mathbf{y}_{t-1}$ will be disappeared, which degenerates to a MLP. 

Taken a Jordan-type RNN as an example, we have:
\begin{align}
\mathbf{\tilde{h}}_t &= f(\mathbf{W^{xh}} \mathbf{\tilde{x}}_t + s_t \mathbf{W^{yh}} \mathbf{y}_{t-1}) \label{Jordan-RNN best} \\
\mathbf{y}_t &= g(\mathbf{W^{hy}} \mathbf{\tilde{h}}_t)
\end{align}
where $\mathbf{\tilde{h}}_t \in \mathbb{R}^{H}$ is the output of the hidden layer. $\mathbf{\tilde{x}}_t$ is the current input. $\mathbf{y}_{t-1} \in \mathbb{R}^{K}$ is the previous output of the output layer. $\mathbf{W^{yh}} \in \mathbb{R}^{H \times K}$ is the recurrent weight from the previous output layer to the current hidden layer.

\subsection{Embedded into Gated RNNs}
For gated RNNs, we use a two-stacked bidirectional LSTM (bi-LSTM) to model the task. The architecture can be defined as follows:
\begin{align} \label{no_skip}
    \overrightarrow{\mathbf{h}_t} &= \text{LSTM}(\overrightarrow{\mathbf{x}_t}, \overrightarrow{\mathbf{h}_{t-1}}, \overrightarrow{\mathbf{c}_{t-1}}) \\
    \overleftarrow{\mathbf{h}_t} &= \text{LSTM}(\overleftarrow{\mathbf{x}_t}, \overleftarrow{\mathbf{h}_{t-1}}, \overleftarrow{\mathbf{c}_{t-1}}) \\
    \mathbf{y}_t &= g(\overrightarrow{\mathbf{h}_t}, \overleftarrow{\mathbf{h}_t})
\end{align}
where $\text{LSTM}(\cdot)$ is the LSTM computation. $\overrightarrow{\mathbf{x}_t}$ and $\overleftarrow{\mathbf{x}_t}$ are the forward and the backward input sequence, respectively. The output of the two hidden layers $\overrightarrow{\mathbf{h}_t}$ and $\overleftarrow{\mathbf{h}_t}$ in a birectional LSTM are stacked on top of each other:
\begin{align}
    \mathbf{h}_t^{l} = f^{l}(\mathbf{h}_t^{l-1}, \mathbf{h}_{t-1}^{l})
\end{align}
where $\mathbf{h}_t^{l}$ is the $t$-th hidden state of the $l$-th layer.

\subsection{Discussion} \label{rel_att}
The main idea of the attention mechanism is to focus on parts of the memories, which are used to store the information for prediction, such as the inputs or the hidden units. From this perspective, our dynamic window method can be seen as an attention-based system. Moreover, supertagging is a special kind of sequence-to-sequence problem, in which the input and the output sequence has the same length. Thus, we do not need to use an encoder to memorize the input and use another decoder to generate the output.

The difference between the two attention mechanisms lies in the type of memories. In the encoder-decoder architecture, the attention model is considered through a weighted average of the output of the encoder. The reason is that they use a encoder and a decoder to model the variable-length outputs, and the memories are the encoder hidden states, while in the supertagging problem, we only use a encoder to do the task, our memories are just the inputs.

\section{Experiments}

We divide our experiments into two steps: First we make comparisons with the existing approaches to test the performance of our models. The comparisons do not include any externally labeled data and POS labels. Then we describe quantitative results which validate the effectiveness of our dynamic window approach. We conduct experiments on MLPs and RNNs with and without this method for comparisons.

\subsection{Dataset and Pre-Processing}
Our experiments are performed on CCGBank \cite{hockenmaier2007ccgbank}, which is a translation from Penn Treebank \cite{marcus1993building} to CCG with a coverage 99.4\%. We follow the standard splits, using sections 02-21 for training, section 00 for development and section 23 for the test. We use a full category set containing 1285 tags. All digits are mapped into the same digit `9', and all words are lowercased.

\subsection{Network Configuration} 
\subsection{Initialization.} There are two types of weights in our experiments: recurrent and non-recurrent weights. For non-recurrent weights, we initialize word embeddings with the pre-trained 200-dimensional GolVe vectors \cite{pennington2014glove}. Other weights are initialized with the Gaussian distribution $\mathcal{N}(0, \frac{1}{\sqrt{\text{fan-in}}})$ scaled by a factor of 0.1, where \textit{fan-in} is the number of units in the input layer. For recurrent weight matrices, we initialize with random orthogonal matrices through SVD \cite{saxe2013exact}to avoid unstable gradients. Orthogonal initialization for recurrent weights is important in our experiments, which takes about $2\%$ relative performance gain than other methods such as Xavier initialization \cite{glorot2010understanding}.

\subsection{Hyperparameters.}
For MLPs, we use a window size of 4, while for vanilla RNNs and gated RNNs, a window size of 1 is enough to capture the local contexts. The dimension of the word embeddings is 200. The size of character embedding and capitalization embeddings are set to 5. We set the maximum number of a word's characters to 5 to eliminate complexity, which means we concatenate the leftmost 5 characters and the rightmost 5 characters as character representations. The number of cells of the stacked bi-LSTM is set to 512. We also tried 400 cells or 600 cells and found this number did not impact performance so much. All stacked hidden layers have the same number of cells. The output layer has 1286 neurons, which equals to the number of tags in the training set with a \textsc{rare} symbol. 

\subsection{Training.}
We train the networks using the back-propagation algorithm, using stochastic gradient descent (SGD) algorithm with an equal learning rate 0.02 for all layers. We also tried other optimization methods, such as momentum \cite{plaut1986experiments}, Adadelta \cite{zeiler2012adadelta}, or Adam \cite{kingma2014adam}, but none of them perform as well as SGD. Gradient clipping is not used. We use on-line learning in our experiments, which means the parameters will be updated on every training sequences, one at a time. 

We use a negative log-likelihood cost to evaluate the performance. Given a training set $\{({\mathbf{x}}^n, \mathbf{t}^n)_{n=1}^N\}$, the objective function can be written as:
\begin{align}
\mathcal{C} = - \frac{1}{N} \sum_{n=1}^N \log {\mathbf{y}}_{t^n}
\end{align}
where $t^n \in \mathbb{N}$ is the true target for sample $n$, and ${\mathbf{y}}_{t^n}$ is the $t$-th output in the \textit{softmax} layer given the inputs $\mathbf{x}^n$.

\begin{figure}
\centering
\includegraphics[width=1.0\linewidth, scale=1.0]{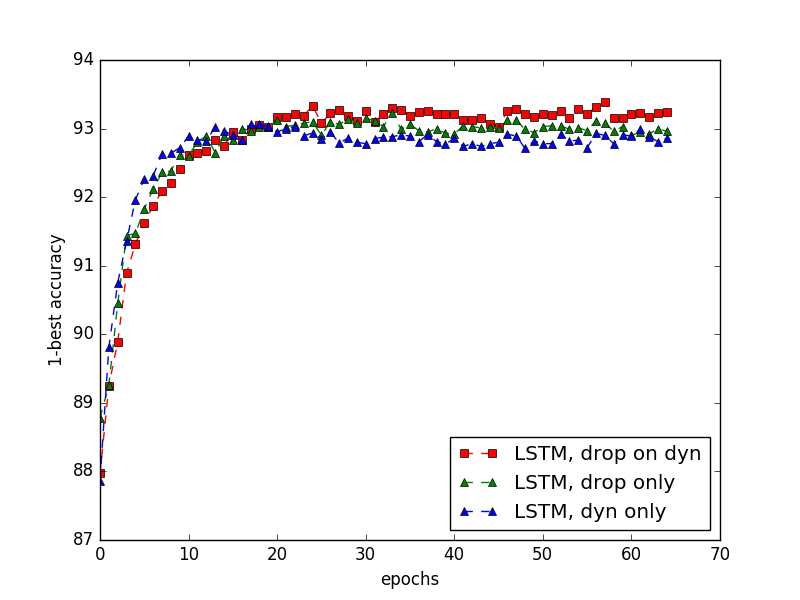}
\caption{The effect of dropout on the filter gate on development set. ``LSTM, drop on dyn'' means the LSTM with dropout on the dynamic window.}
\label{fig:dropout}
\end{figure}

\subsection{Regularization.}
Dropout is the only regularizer in our model to avoid overfitting. We add a dropout mask to our filter gate $\mathbf{r}_t$ with a drop rate 0.5, which is helpful to improve the performance. Figure \ref{fig:dropout} shows such comparisons on a forward LSTM. The behavior of other kinds of neural networks are similar. We can see that dropout on the words directly Eq. \eqref{eq:one} is slightly better than the dynamic window without dropout Eq. \eqref{eq:filter-lt}, while dropout on the dynamic window lead to a much better improvement.  We also apply dropout to the output of the hidden layer with a 0.5 drop rate. At test time, weights are scaled with a factor $1-p$.

\subsection{Results on Supertagging Accuracy}
We report the highest 1-best supertagging accuracy on the development set for final testing. Table \ref{CCGBank} shows the comparisons of accuracy on CCGBank. We notice that stacked bi-LSTM (with depth 2) performs the best than other neural based models. Our dynamic window approach provides the highest (+11\%) relative performance gain. The character-level information (+ 6\% relative accuracy) and dropout (+ 8\% relative accuracy) are also helpful to improve the performance. We observe that dropout on words is superior to dropout on word embeddings.

\begin{table}[h]
\begin{center}
\begin{tabular}{ l | l | l }
\hline Model & Dev & Test \\ \hline \hline
Clark and Curran \shortcite{clark2007wide} & 91.5 & 92.0 \\
Lewis et al. \shortcite{lewis2014improved}  & 91.3 & 91.6 \\
Lewis et al. \shortcite{lewis2016lstm} & 94.1 & 94.3 \\
Xu et al. \shortcite{xu2015ccg} & 93.1 & 93.0 \\
Xu et al. \shortcite{xu2016expected} & 93.49 & 93.52 \\
Vaswani et al. \shortcite{vaswani2016supertagging} &  94.24 & 94.5 \\ 
\hline
MLP & 92.06 & 92.28 \\
Elman RNN & 92.74 & 92.89 \\
Jordan RNN & 92.61 & 92.75 \\ 
forward LSTM & 93.39 & 93.51 \\
bi-LSTM & 94.1 & 94.13 \\
stacked bi-LSTM (depth 2) & \bf 94.4 & \bf 94.69 \\
stacked bi-LSTM (no dyn) & 94.02 & 94.09 \\
stacked bi-LSTM (no char) & 93.89 & 94.21 \\
stacked bi-LSTM (drop emb) & 94.13 & 94.33 \\
stacked bi-LSTM (no dropout) & 94.06 & 94.25 \\
\hline
\end{tabular}
\end{center}
\caption{\label{CCGBank} 1-best supertagging accuracy on CCGbank. ``no dyn'' refers to the models that do not use the filter gates to concatenate the tokens in a context window, ``no char'' refers to the models that do not use the character-level information, ``drop emb'' refers to dropout on word embeddings rather than on dynamic filters.}
\end{table}

\subsection{On the Usage of Dynamic Filters}
We experiment with a $1 \times 1$ convolution operation on $\mathbf{x}_t$. The performance is 94.05\% (Table \ref{tab-filters}, line 2), which indicates that the convolution is recommended to operate on words directly, rather than on word embeddings. This can be formulated as:
\begin{align}
&\mathbf{\tilde{x}}_t = \mathbf{r}_t \odot \mathbf{x}_t \\ 
&:= [\mathbf{r}_{{t-\lfloor d/2 \rfloor}} \odot \mathbf{f}_{w_{t-\lfloor d/2 \rfloor}}; \ldots  ; \mathbf{r}_{{t+\lfloor d/2 \rfloor}} \odot \mathbf{f}_{w_{t+\lfloor d/2 \rfloor}}]
\end{align}
Here $\odot$ denote an element-wise product. 

We can use a MLP instead of a one-layer network to learn the dynamic filters:
\begin{align}
    \mathbf{u}_t &= \sigma(\mathbf{W^{xu}} \mathbf{x}_t) \\
	\mathbf{r}_t &= \sigma(\mathbf{W^{ur}} \mathbf{u}_t)
\end{align}
where we add a hidden layer $\mathbf{u}_t \in \mathbb{R}^u$ to learn $\mathbf{r}_t$. But the performance is 94.23\% (Table \ref{tab-filters}, line 3) with an extra computational cost.

The weighted average on the inputs $\sum_{i=1}^d r_i \mathbf{f_i}$ is a standard method for attention, which leads to a poor result of 93.95\% (Table \ref{tab-filters}, line 4). This shows the gated concatenation might be superior to the weighted average in the context of sequence tagging.

\begin{table}[t]
\begin{center}
\begin{tabular}{l | l | l}
\hline 
Filters & Accuracy & Remark \\
\hline \hline
one layer (origin) & 94.4 & $\mathbf{r}_t \in \mathbb{R}^d$, $\mathbf{\tilde{x}}_t = \mathbf{r}_t \otimes \mathbf{x}_t$ \\ \hline
one layer (CNN) & 94.05 & $\mathbf{r}_t \in \mathbb{R}^I$, $\mathbf{\tilde{x}}_t = \mathbf{r}_t \odot \mathbf{x}_t$ \\ \hline
two layer(MLP) & 94.23 & \\ \hline
averaging & 93.95 & $\mathbf{\tilde{x}}_t = \sum_{i=1}^d r_i \mathbf{f_i}$ \\ \hline
\end{tabular}
\end{center}
\caption{\label{tab-filters} 1-best tagging accuracy on the development set with different dynamic filters, using a stacked bi-LSTM model. ``two layer'' refers to using a two layer feed-forward neural network to learn dynamic filters. $\mathbf{r}_t \odot \mathbf{x}_t$ is a element-wise product.}
\end{table}

\subsection{Effects of the Dynamic Window Approach}
\begin{figure}[h]
\centering
\includegraphics[width=1.0\linewidth, scale=1.0]{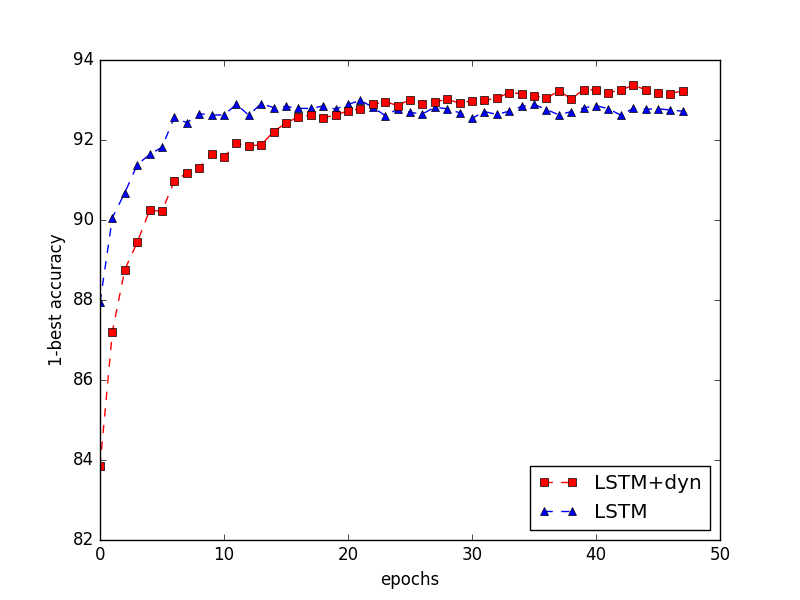}
\caption{1-best accuracy of RNN(LSTM) models with and without the dynamic window approach on the development set.}
\label{fig-dyn}
\end{figure}

Figure \ref{fig-dyn} shows the effectiveness of our dynamic window approach. Taken a forward LSTM as an example, we can observe at first 20 epochs the LSTM + dyn performs worse than the original LSTM model since many useful input contexts are filtered using this approach, but after 20 epochs the LSTM starts to overfitting while the performance of the LSTM + dyn continues to raise up.

\begin{figure*}
\centering
\begin{subfigure}{.3\textwidth}
  \centering
  \includegraphics[width=.6\linewidth]{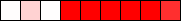}
  \caption{$C(\text{make})$}
  \label{fig:tfig1}
\end{subfigure}%
\begin{subfigure}{.3\textwidth}
  \centering
  \includegraphics[width=.6\linewidth]{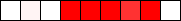}
  \caption{$C(\text{series})$}
  \label{fig:tfig2}
\end{subfigure}
\begin{subfigure}{.3\textwidth}
  \centering
  \includegraphics[width=.6\linewidth]{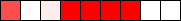}
  \caption{$C(\text{chance})$}
  \label{fig:tfig3}
\end{subfigure} 

\begin{subfigure}{.3\textwidth}
  \centering
  \includegraphics[width=.6\linewidth]{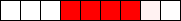}
  \caption{$C(\text{the})$}
  \label{fig:tfig4}
\end{subfigure}%
\begin{subfigure}{.3\textwidth}
  \centering
  \includegraphics[width=.6\linewidth]{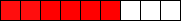}
  \caption{$C(\text{review})$}
  \label{fig:tfig5}
\end{subfigure}
\begin{subfigure}{.3\textwidth}
  \centering
  \includegraphics[width=.6\linewidth]{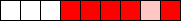}
  \caption{$C(\text{he})$}
  \label{fig:tfig6}
\end{subfigure}
\caption{Examples of the activations of filter gates for different contexts. (\emph{red} indicates filters saturated near 1, \emph{white} indicates filters unsaturated near 0), $C(\cdot)$ refers to the fixed context surrounding the center word.}
\label{fig:vis_cont}
\end{figure*}

\subsection{Visualizations}
Our model uses filter gates to dynamically choose the needed contexts. To understand this mechanism, we randomly choose some words to visualize the dynamic activities during training. All the visualizations are done using an MLP on CCGBank Section 02-21.

Figure \ref{fig:vis_cont} shows the different dynamic activities for the words. Each sub-figure has 9 blocks (a window size of 4), each of which shows an activation of one filter $l_i(t), i \in \{1, \ldots, 9\}$. After training to convergence, the items far away from the middle words are gradually removed from the contexts, while the attentions are gradually focused on the items nearby the central words. We can observe that for different words, their dynamic activities are different.

\section{Related Work}

Clark and Curran \shortcite{clark2007wide} use a log-linear model to build the supertagger, using discrete feature functions for the targets based on the words and POS tags. The discrete property of the model makes the features independent of each other. Lewis and Steedman \shortcite{lewis2014improved} propose a semi-supervised supertagging model using a multi-layer perceptron (MLP) based on Collobert \shortcite{collobert2011natural} and conditional random field (CRF) proposed by Turian et al. \shortcite{turian2010word}. Without using POS tags, they use the per-trained word embeddings with 2-character suffix and capitalization as features to represent the word. This distributed embedding encodes the word similarities and provides a better representation than log-linear models. However, MLP based supertagger ignores the sequential information, and their CRF based model can capture this but takes far more computational complexity than the MLP model due to the huge number of supertags.

The supertaggers based on log-linear and MLP are all point estimators, while CCG supertagging is more suitable to be treated as a sequential estimation problem due to long-range dependencies of the predicate-argument relations contained in lexical categories. Recently, Xu et al. \shortcite{xu2015ccg} design an Elman-type RNN to capture these dependencies, and use a fixed size window for each word as MLPs. The recurrent matrix in RNN can restore the historical information, which makes it outperform the MLP based model. But RNNs may suffer from the gradient vanishing/exploding problems and are not good at capturing long-range dependencies in practice. Vaswani et al. \shortcite{vaswani2016supertagging} and Lewis et al. \shortcite{lewis2016lstm} shows the effectiveness of bi-LSTMs in supertagging, but they do not use a context window for the inputs. We only get 93.9\% performance on the development set without using context windows. We find that a window size of 1 is needed in the stacked bi-LSTM to get a better performance.

Our model can be treated as a marriage between attention mechanism and dropout. The most relevant attention-based models relating to our work is Wang et al. \shortcite{wang:2015-2}, in which they use an attention model to find the relevant words within the context for predicting the center word. Their attention mechanism is similar to Bahdanau et al. \shortcite{bahdanau2014neural}, while ours was not originally designed as a weighted average but a gated concatenation. Dropout on the dynamic window is similar to \cite{dai2015semi}, which randomly drop words in the input sentences. Gal \shortcite{gal2015theoretically} also use dropout on words, but using a fixed mask rather a random one.

\section{Conclusion}
We presented a dynamic window approach for CCG supertagging. Our model uses logistic gates to filter the context window surrounding the center word. This attention mechanism shows effectiveness on both MLPs and RNNs. We observed that using dropout on the dynamic window will greatly improve the generalization performance. We further visualized the activation of the filters, which is useful to help us understanding the dynamic activities. Although our work mainly focus on the CCG supertagging, this method can be easily applied to other sequence tagging tasks, such as POS tagging and named entity recognition (NER).



\bibliographystyle{acl2012}
\bibliography{acl2012}
\end{document}